\documentclass[]{mosi}

\usepackage[T1]{fontenc}
\usepackage{enumitem}
\usepackage{amsmath,amssymb,amsfonts}
\usepackage{natbib}
\usepackage{graphicx}
\usepackage{subcaption}
\usepackage{booktabs}
\usepackage{multirow}
\usepackage{makecell}
\usepackage{array}
\usepackage{threeparttable}

\usepackage{xcolor}
\usepackage{float}
\usepackage{setspace}
\usepackage{nicefrac}

\usepackage[T1]{fontenc}
\usepackage{enumitem}
\usepackage{amsmath,amssymb,amsfonts}
\usepackage{natbib}
\usepackage{graphicx}
\usepackage{subcaption}
\usepackage{booktabs}
\usepackage{multirow}
\usepackage{makecell}
\usepackage{array}
\usepackage{threeparttable}

\usepackage{xcolor}
\usepackage{float}
\usepackage{setspace}
\usepackage{nicefrac}
\usepackage{microtype}

\usepackage[toc,page,header]{appendix}
\usepackage{hyperref}
\usepackage{url}
\usepackage{wrapfig}

\addto\extrasenglish{

}

\title{HiMe: Hierarchical Embodied Memory for Long-Horizon Vision-Language-Action Control}

\newcommand{\equal}{*}
\newcommand{\corr}{\ensuremath{\dagger}}
\newcommand{\pjlead}{\ensuremath{\ddagger}}

\author{
Li Ji$^{1,2,\equal}$ \hspace{.3em}
Siyin Wang$^{1,2,\equal,\pjlead}$ \hspace{.3em}
Pengfang Qian$^{1,2}$ \hspace{.3em}
Xiaopeng Yu$^{1}$ \hspace{.3em}
Yihai Tian$^{2,3}$ \hspace{.1em}
\newline
Zhaoye Fei$^{1}$ \hspace{.1em}
Jingjing Gong$^{2,\corr}$ \hspace{.1em}
Xipeng Qiu$^{1,2,\corr}$
\\[1ex]

\texttt{
$^{1}$Fudan University   
$^{2}$Shanghai Innovation Institute   
$^{3}$East China Normal University 
}
\newline
{\fontsize{10}{12}\selectfont \texttt{$^*$Equal Contribution  $^{\ddagger}$Project Lead $^{\dagger}$Corresponding Author}}
\newline
{\fontsize{10}{12}\selectfont \texttt{24210240184@m.fudan.edu.cn, siyinwang20@fudan.edu.cn}}
}

\abstract{
Current Vision-Language-Action (VLA) models excel at robotic manipulation but often struggle with non-Markovian tasks requiring long-term memory and reasoning due to their reliance on \mbox{immediate} observations. Existing solutions  face a ``frequency-competence paradox,'' where stronger reasoning models are too slow for real-time control, while faster models lack sufficient reasoning capabilities. To resolve this architectural misalignment, we propose \textbf{HiMe}, a Hierarchical Embodied Memory \mbox{framework} that decouples embodied intelligence into a high-frequency Executor for execution, a Sentry for working memory, and a Planner for long-term strategy. We also introduce a dynamic knowledge system based on cross-modal semantic schemas and active management mechanisms, \mbox{allowing} robots to maintain memory plasticity through ``Add, Update, and Delete'' operations. This hierarchical design effectively balances the conflict between real-time execution and slow \mbox{thinking} \mbox{planning}, significantly improving success rates in long-horizon tasks. Experiments demonstrate that this \mbox{approach} not only outperforms flat memory baselines but also exhibits the novel ability to \mbox{self-correct} its internal knowledge based on human preferences.
}

\checkdata[Code]{\url{https://github.com/HappyWaterXP/HiMe}}

\begin{document}
\maketitle

\begingroup
\renewcommand{\thefootnote}{\fnsymbol{footnote}}
\setcounter{footnote}{1}
\endgroup

\section{Introduction}

Vision-Language-Action (VLA) models have emerged as a powerful paradigm for general-purpose robotic manipulation with large-scale internet-level pretraining \cite{openvla,Intelligence202505AV}. By directly mapping observations to control signals, they provide a powerful foundation for executing complex motor skills. 
Most existing architectures rely on the Markov assumption, where the policy $p(a_t | o_t, l)$ predicts the action $a_t$ at time step $t$ conditioned only on the transient observation $o_t$ at the current time step and the language instruction $l$. This inherent limitation prevents them from maintaining a persistent belief of the environment in non-Markovian settings. 
Consequently, VLAs struggle with complex, long-horizon tasks that demand long-range temporal dependencies, where the optimal action depends on a chain of past events or latent information that is no longer visible in the current sensory stream.

\begin{figure}[H]
    \centering
    \includegraphics[width=1\linewidth]{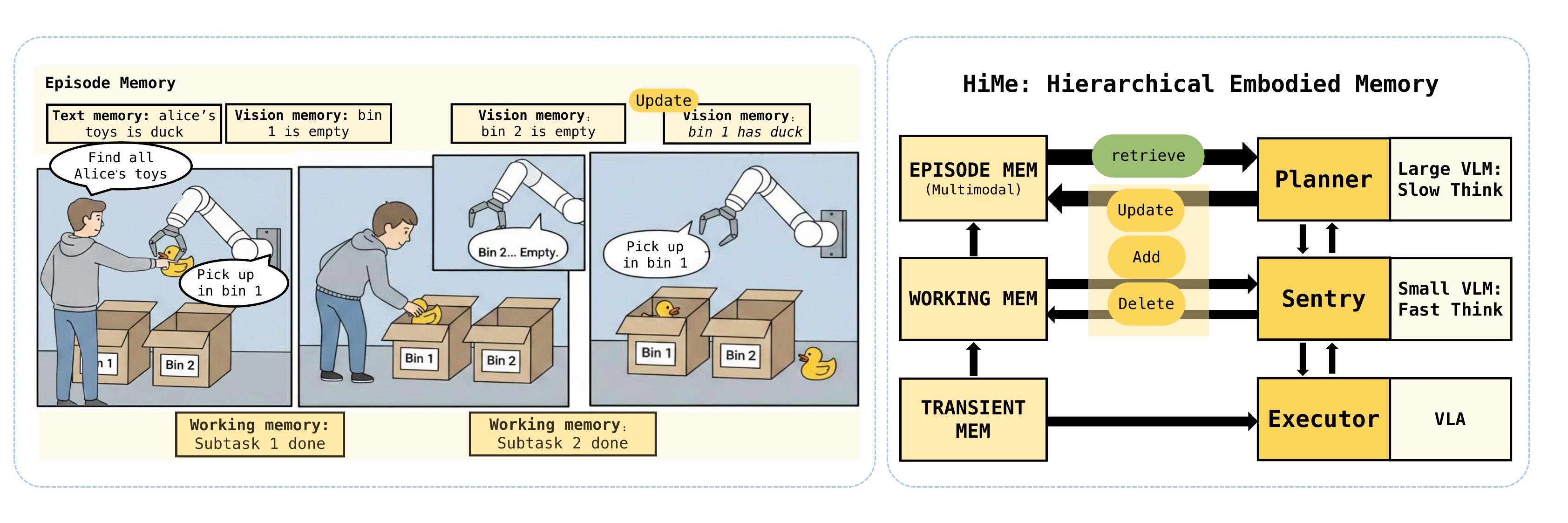}
    \caption{\textbf{Overview of HiMe.} (a) A motivating example illustrating the need to maintain and update task-relevant multimodal memory across long-horizon subtasks. (b) The HiMe framework, which addresses this challenge by organizing embodied intelligence into a hierarchical structure that separates fast execution from memory-driven reasoning over long time scales.}
    \label{fig:intro}
\end{figure}

To overcome these limitations, one intuitive approach is to imbue VLA models with native memory capabilities through specialized training or auxiliary losses \cite{Torne2025LearningLD,Fang2025SAM2ActIV}. However, these training-based methods are often constrained by limited context windows and the inherent difficulty of optimizing long-range causal dependencies over hundreds of steps. Another alternative is to introduce Large Vision-Language Models (LVLMs) as high-level memory containers to store and retrieve historical trajectories \cite{Sridhar2025MemERSU}. Yet, real-time robotic control imposes a strict upper bound on the deployable VLM scale, as control demands high-frequency, low-latency execution. This scale constraint, in turn, limits the internal world knowledge and generalization capabilities of the VLM, weakening its zero-shot performance. Importantly, most memory operations do not intrinsically require high-level reasoning or broad generalization. 


As illustrated in~\autoref{fig:intro}, motivated by this temporal and scale mismatch, we introduce \textbf{HiMe}, a \textbf{Hierarchical Embodied Memory} framework that decouples embodied intelligence into three functional layers with distinct temporal resolutions, mirroring the multi-store structure of human cognition. 
(1) The \textbf{Executor (VLA)} governs \textbf{transient memory}, focusing on high-frequency sensory-motor coordination for physical stability. 
(2) The \textbf{Sentry} acts as the guardian of \textbf{working memory} (short-term memory); it asynchronously filters the continuous sensory stream to identify critical state transitions, effectively bridging the gap between raw perception and semantic events. 
(3) The \textbf{Planner} manages \textbf{episodic memory} (long-term memory). 
By invoking the ``slow-thinking" Planner only at the discrete junctions identified by the Sentry, our architecture preserves deep strategic reasoning while maintaining real-time execution, effectively resolving the frequency-capability paradox.

However, a static memory hierarchy alone is insufficient for the complexities of real-world human-robot interaction, which is characterized by: (1) \textit{multimodal richness}, where human instructions carry dense logical constraints and latent preferences \cite{roboomni}, such as designating a specific red cup as ``Alice's cup'', that are nearly impossible to capture through vision-only trajectories; and (2) \textit{high dynamism}, where task goals and environment states are not static but evolve. A memory system that only supports passive accumulation~\cite{Sridhar2025MemERSU} inevitably suffers from knowledge stagnation and cognitive dissonance when faced with such outdated or conflicting experiences.

Furthermore, our Hierarchical Memory framework treats memory as a multimodal, dynamic knowledge system centered around two fundamental dimensions:

\textit{(1) What to memorize?: Cross-modal Semantic Schemata.} To bridge the gap between raw perception and logical intent, we represent episodic memory as an object-centric system where visual experiences are deeply anchored with high-density textual descriptions. This organization allows the Planner to retrieve not only visual features but also the underlying ownerships, procedural rules, and human preferences that are invisible in pure pixel space.

\textit{(2) How to memorize?: Active Management Mechanism.} In contrast to passive storage, we introduce explicit \textit{Add}, \textit{Update}, and \textit{Delete} operations to grant the robot knowledge plasticity. When the Sentry detects a significant state transition or a revision in user intent, the Planner proactively refines the knowledge base. By updating outdated beliefs and purging redundant or erroneous data, the agent maintains a consistent and concise memory that evolves in alignment with the environment.

We evaluate our method across a suite of challenging long-horizon manipulation tasks that require multi-step reasoning and adaptation to shifting human preferences. Experimental results show that our approach significantly outperforms existing flat-memory baselines in both success rate and computational efficiency. Notably, our agent demonstrates the ability to ``self-correct" its internal knowledge base when faced with conflicting instructions, a capability that is absent in prior retrieval-based methods.

Our core contributions are summarized as follows:
\begin{itemize} 
    \item We propose a \textbf{Hierarchical Memory Management framework} that decouples robotic control into transient (Executor), working (Sentry), and episodic (Planner) memory layers, resolving the granularity conflict in long-horizon VLA tasks.
    \item  We introduce a Cross-modal Memory organization (\textit{what to memorize}) combined with an Active Management mechanism (\textit{how to memorize}), enabling the robot to maintain knowledge plasticity and alignment in highly dynamic and multimodal interactions.
    \item We provide \textbf{extensive empirical evidence} demonstrating that our hierarchical, self-evolving memory architecture achieves superior performance and robustness in complex human-robot collaborative scenarios with significantly reduced VLM computational overhead.
\end{itemize}

\section{Related Work}
\subsection{Foundation models in robotics}
Recent advancements in End-to-End Vision-Language-Action (VLA) models adopt a co-training paradigm that integrates a pre-trained VLM backbone with a dedicated action expert~\cite{Intelligence202505AV}. By harnessing the VLM's extensive world knowledge~\cite{Bai2025Qwen3VLTR} to interpret complex scenes, this approach effectively transforms the backbone into a generalist agent capable of mapping raw observations directly to continuous actions. 
Alternatively, the hierarchical paradigm decouples reasoning from execution~\cite{Shi2025HiRO, Li2025HAMSTERHA, Shentu2024FromLT}. In this framework, a high-level policy, typically a VLM, processes visual observations and task descriptions to generate intermediate goals, such as natural language sub-tasks or latent embeddings. These intermediate representations then condition a low-level policy to output continuous actions. Building upon this hierarchical structure, our work introduces a novel hierarchical memory architecture designed to empower the high-level VLM to generate more precise and context-aware instructions for the low-level controller.

\subsection{Robotic memory}

Memory is essential for robotic agents operating in long-horizon, partially observed manipulation tasks, where successful execution depends on retaining past observations, inferred world states, and user-specific context. Prior work has explored different mechanisms for incorporating historical information into robotic control. One major direction introduces memory at the policy level, primarily to improve local action generation, spatial grounding, and short-range temporal consistency. These approaches either augment Vision-Language-Action models with retrieval-style memory banks that store and retrieve relevant past context~\cite{Li2025MAPVLAMP, Fang2025SAM2ActIV, Zhu2020VisionDialogNB, Shi2025MemoryVLAPM}, or encode history into compact representations such as past tokens or visual traces for history-conditioned action prediction~\cite{Torne2025LearningLD, Zheng2024TraceVLAVT}.

Another line of work studies memory in more structured, higher-level decision processes. Hierarchical embodied-agent systems~\cite{Wang2023JARVIS1OM, Li2024Optimus1HM, Wang2023DescribeEP} highlight the importance of explicit planning, reflection, and multimodal memory for long-horizon reasoning. MemER~\cite{Sridhar2025MemERSU} further organizes experience using a FIFO queue with keyframe selection, demonstrating that memory supports not only short-range control but also the maintenance of task-level context over extended horizons. Our work follows this direction and focuses on a multimodal memory design with explicit memory management.

\subsection{Memory management in Agents}
Effective memory management is crucial for handling long-context interactions. 
Existing frameworks typically organize memory using hierarchical structures~\cite{Liu2025MemVerseMM} or OS-inspired interfaces~\cite{Packer2023MemGPTTL}, often relying on explicit management strategies~\cite{Chhikara2025Mem0BP, Xu2025AMEMAM, Yan2025MemoryR1EL} to maximize efficiency.
A dominant strategy among these is periodic summarization~\cite{Long2025SeeingLR, Yeo2025WorldMMDM}, where recent history is compressed into long-term storage at fixed temporal intervals to decouple storage from reasoning.
However, these methods have predominantly focused on text-based representations for LLMs~\cite{Liang2023SCMEL}.
With the rapid evolution of Vision-Language Models (VLMs), there is a growing necessity to incorporate multimodal information.
Our approach addresses this by constructing a cross-modal memory that synergizes text and images, while simultaneously introducing a structured management mechanism tailored to effectively utilize this rich context for complex robotic tasks.

\section{Methods}
\begin{figure*}
    \centering
    \includegraphics[width=0.92\linewidth]{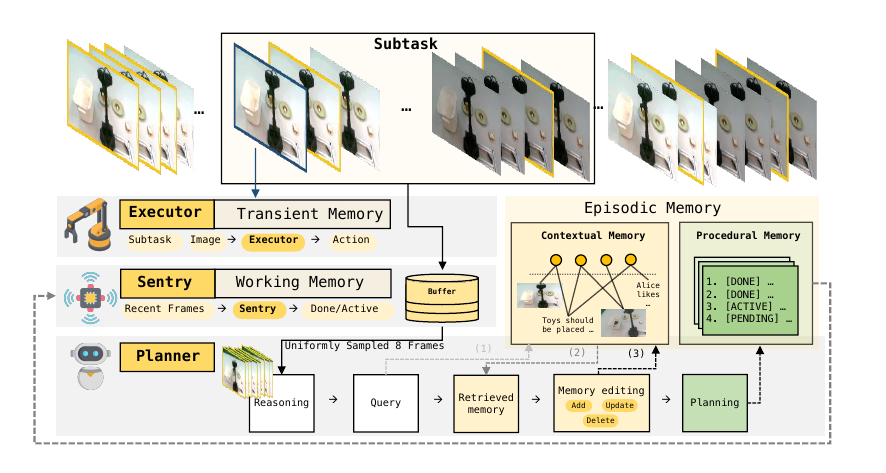}
    \caption{\textbf{Architecture of HiMe.} The Sentry module serves as a real-time monitor, periodically tracking the Executor's progress and buffering visual observations. Upon detecting the completion of a subtask, it triggers the Planner to review the entire execution trajectory. The Planner then performs memory retrieval and consolidation to update the contextual memory and refines the procedure memory. Finally, the refined instruction is fed back to the Sentry to resume monitoring the Executor, forming a robust closed-loop system. }
    \label{fig:hime}
\end{figure*}

\subsection{Overview}
Long-horizon manipulation relies on two contradictory capabilities: maintaining global consistency over extended history and reacting responsively to immediate dynamics. End-to-end approaches typically struggle to balance these needs—short-context policies suffer from catastrophic forgetting, while heavy reasoning models incur prohibitive latency for real-time control.

To address this challenge, we propose HiMe, a Hierarchical Embodied Memory framework (\autoref{fig:hime}) that decouples embodied intelligence into three functional layers: 
(1) The Executor $\pi_e$: A high-frequency VLA that ensures physical stability by mapping immediate observations to actions. 
(2) The Sentry $\pi_s$: A lightweight VLM that identifies progress changes in the environment, triggering planning only when necessary.
(3) The Planner $\pi_p$: A heavyweight VLM that uses long-term episodic memory and plans at a lower frequency.
This separation amortizes the cost of complex planning over extended execution windows, ensuring the low-level controller remains responsive to immediate physical dynamics while maintaining long-term task coherence. 

\textbf{Problem Setup.} 
We formulate the long-horizon manipulation task as a sequential decision-making process under partial observability. The agent operates over a horizon $T \gg 1$, receiving high-dimensional observations $o_t \in \mathcal{O}$ and a natural language instruction $l$. The objective is to generate an action sequence $a_{0:T}$ that transitions the environment to a state satisfying the instruction $l$. Ideally, the optimal action at any step $t$ depends on the full interaction history $h_t = (o_0, a_0, \dots, o_t)$ to resolve ambiguity and track progress. However, processing $h_t$ continuously is computationally intractable. To address this, we adopt a hierarchical policy $\pi\!\left(a_t \mid o_{0:t}, l\right)$.

\begin{figure*}[t]
    \centering
    \includegraphics[width=1\linewidth]{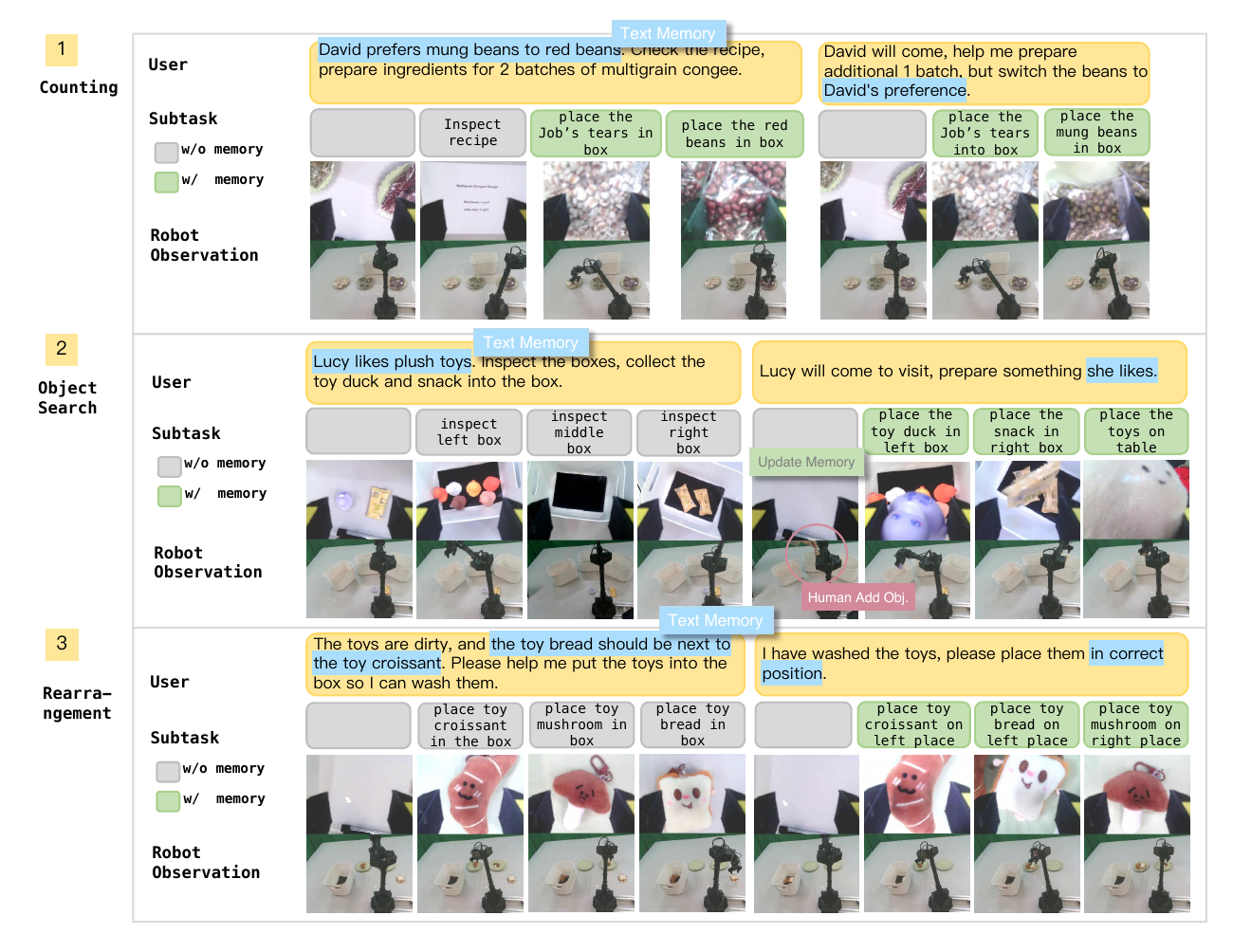}
    \caption{Tasks used in our evaluations. We report performance across 20 trials per task per method.}
    \label{fig:exp}
\end{figure*}

\subsection{Hierarchical Embodied Memory}
To bridge the gap between reactive control and reflective reasoning, we augment the agent with a structured state space comprising three distinct components. This explicitly represents the information required for immediate control, temporal monitoring, and long-term planning.

\textbf{Transient Memory ($\mathcal{T}_t$).} Managed by the Executor, it contains the instantaneous observation $o_t$. It serves as the minimal sufficient state for reactive sensory-motor coordination.

\textbf{Working Memory ($\mathcal{W}_t$).} 
To support high-frequency monitoring, we maintain a sliding window of recent observations $\mathcal{W}_t = \{o_{t-h_s}, \dots, o_t\}$. Unlike the main memory, this buffer is transient and raw, capturing the immediate dynamics required for verifying subgoal status or detecting sudden failures.

\textbf{Episodic Memory ($\mathcal{E}_t$).} A persistent long-term store managed by the Planner, consisting of: 
(1) \textit{Contextual Memory $\mathcal{E}_t^{c}$} 
, a multimodal key--value store $\mathcal{M}_t$ that accumulates compact, task-relevant information—including both visual imagery and textual descriptions—over time, such as object state changes, spatial constraints, or human preferences, and supports retrieval via textual keys.
(2) \textit{Procedural Memory $\mathcal{E}_t^{p}$} 
, a structured plan $\mathcal{E}_t^{p} = \{\tau_i\}_{i=1}^{N}$ consisting of an ordered list of language subgoals, where the active subgoal $\tau_t^{(i)}$ carries status labels like `active', `pending' or `done' to synchronize the control loop.


\subsection{Hierarchical Policy Decomposition}
We decompose the hierarchical policy $\pi\!\left(a_t \mid o_{0:t}, l\right)$ into a three-tiered hierarchy. 
This design ranges from a memory-free execution layer to a long-horizon reasoning layer, progressively trading off frequency for context capacity, decoupling immediate actuation from high-level reasoning. Formally, the decision process at time $t$ involves interactions between a base Executor $\pi_e$, a monitoring Sentry $\pi_s$, and a reasoning Planner $\pi_p$.

\textbf{Base-level Executor ($\pi_e$).}
At the base level, the Executor functions as a high-frequency reactive policy. We assume a local Markov property where, given an active language subgoal $\tau_t \in \mathcal{E}_t^p$, the optimal action is conditionally independent of the long-term history. The Executor maps the current observation and subgoal to an action:
\begin{equation}
    a_t \sim \pi_e\!\left(a_t \mid o_t, \tau_t\right),
\end{equation}
where $\tau_t$ serves as a conditioning variable abstracted from the complex episodic history, representing the current executing subgoal. This design ensures that $\pi_e$ remains stateless and computationally efficient, enabling real-time responsiveness to physical dynamics.

\textbf{Agile Sentry ($\pi_s$).}
Consistent with the architecture in \autoref{fig:hime}, the Sentry acts as a gating function $u_t \in \{0, 1\}$ that is used to amortize the computational cost of reasoning. Operating on the sliding window working memory $\mathcal{W}_t$, the Sentry evaluates subgoal completion or environmental deviation at intervals of $n_m$:
\begin{equation}
u_t =
\begin{cases}
\mathbf{I}\!\left[\pi_s\!\left(\mathcal{W}_t,\tau_t\right) > \delta\right],
& \text{if } t \equiv 0 \; (\mathrm{mod}\; n_m),\\
0, & \text{otherwise},
\end{cases}
\end{equation}
where $\mathbf{I}[\cdot]$ is the indicator function and $\delta$ is a decision threshold. $u_t=1$ signals a ``handover" event—indicating that the current subgoal $\tau_t$ is either completed or invalidated, thereby triggering the Planner. When $u_t=0$, the system bypasses high-level reasoning, maintaining the current episodic memory $\mathcal{E}_{t\!+\!1} \leftarrow \mathcal{E}_t$.

\textbf{Reasoning Planner ($\pi_p$).}
When the Sentry trigger is activated ($u_t = 1$), the Planner is invoked to update the episodic memory and re-align the agent's internal belief with the current environment. Let $l$ denote the user instruction, $\mathcal{W}_t = \{o_{t-h_s}, \dots, o_t\}$ the working memory containing recent observations, and $\mathcal{E}_t = \left(\mathcal{E}_t^c, \mathcal{E}_t^p\right)$ the episodic memory maintained by the Planner, where $\mathcal{E}_t^c$ stores contextual information and $\mathcal{E}_t^p$ represents the current procedural plan.

This update proceeds in two stages.

(1) \textit{Context Retrieval:}
The Planner first encodes the instruction and recent observations into a query, and retrieves the most relevant contextual entries from memory:
\begin{equation}
    q_t = f_{\text{enc}}\!\left(l, \mathcal{W}_t\right), \qquad
    \mathcal{M}_{\mathrm{ret}} = \mathrm{TopK}\!\left(\mathcal{E}_t^c, q_t\right),
\end{equation}
where $\mathcal{M}_{\mathrm{ret}} \subset \mathcal{E}_t^c$ denotes the retrieved subset of contextual memory.

(2) \textit{Memory Update and Re-planning:}
Conditioned on the instruction $l$, working memory $\mathcal{W}_t$, retrieved context $\mathcal{M}_{\mathrm{ret}}$, and the current procedural plan $\mathcal{E}_t^p$, the Planner jointly updates both components of episodic memory:
\begin{equation}
    \left(\mathcal{E}_{t+1}^c, \mathcal{E}_{t+1}^p\right)
    \leftarrow
    \pi_p\!\left(l, \mathcal{W}_t, \mathcal{M}_{\mathrm{ret}}, \mathcal{E}_t^p\right).
\end{equation}
Specifically, the update to contextual memory $\mathcal{E}^c$ is realized through three explicit operations: \texttt{Add}, which inserts newly observed facts or user preferences; \texttt{Update}, which revises outdated entries; and \texttt{Delete}, which removes stale or conflicting information. In parallel, the Planner synthesizes an updated procedural plan $\mathcal{E}_{t+1}^p$, from which the next active subgoal $\tau_{t+1}$ is selected to continue execution.

\subsection{Implementation Details} 
While our framework is model-agnostic, we instantiate the Planner with GPT-4o~\cite{Hurst2024GPT4oSC} for its strong multimodal reasoning capabilities, and the Sentry with Qwen3-VL-8B~\cite{Bai2025Qwen3VLTR} for its balance of speed and grounding accuracy. The memory backend utilizes a vector database for semantic retrieval, utilizing OpenAI's text-embedding-3 model to encode text queries and storing multimodal memory entries alongside their caption embeddings for cosine similarity retrieval. We set the buffer size $h_s = 8$ and the monitoring interval $n_m = 5$. Consistent with our design philosophy, only the lightweight $\pi_e$, a fine-tuned VLA based on $\pi_{0.5}$, requires domain-specific training. The Sentry and Planner operate in a zero-shot or few-shot manner, leveraging pre-trained generalization to handle long-horizon logic without extensive data collection.

\section{Experiments}
\subsection{Task Setting}

To demonstrate the critical role of hierarchical memory in long-horizon tasks, we designed three distinct tabletop manipulation scenarios, as illustrated in~\autoref{fig:exp}. These tasks are specifically curated to evaluate diverse capabilities, including user interaction, preference memory, updating memory and planning with exploration.

\textbf{Object Search.} 
We design a Domestic Maintenance task to evaluate the robot's capability to integrate inspection, sorting, and preference recall. In this scenario, the robot must first inspect opaque boxes to deduce storage rules and organize scattered items accordingly. Crucially, during this process, the user introduces new toys into the boxes to test the robot's ability to dynamically update its memory. Finally, the robot faces a retrieval challenge, where it must synthesize memorized user preferences with the box contents identified during inspection to retrieve the correct object.

\textbf{Counting.} 
In this scenario, the robot is required to interpret a visual recipe on the table and plan its subsequent actions. The task proceeds in stages: the robot must first \textit{inspect} the recipe to extract the ingredient composition and prepare the materials based on the number of servings requested by the user. Following this, the user introduces a specific preference. To fulfill this customized demand, the robot must synthesize the new constraint with the previously memorized recipe to dispense the correct personalized ingredients.

\textbf{Rearrangement.} 
In this playroom scenario, the robot performs a two-stage ``clear and restore'' operation. Initially, the robot collects all toys scattered on the table into a storage box for cleaning. After a temporal interval, the robot is tasked with restoring the items to the environment. Crucially, the user's specific placement preference is provided beforehand as prior knowledge. Therefore, to execute the restoration, the robot must recall this pre-established preference and synthesize it with its visual memory of the original spatial configuration to plan the final placement of each item.

\textbf{Evaluation Setup.}
We use a WidowX-250s arm with a parallel gripper and dual-camera visual input (third-person and wrist view). The Executor $\pi_e$ runs at 2\,Hz, predicting an action chunk $A_t$ of 10 actions (10\,Hz), of which 5 are executed open-loop. To monitor the subtask progress, the Sentry $\pi_s$ is queried after every 10 execution steps of $\pi_e$. The Sentry will trigger Planner $\pi_p$ if the current subtask is completed to invoke high-level planning.
Each task’s evaluation metrics are detailed in \autoref{tab:task_metrics}.




\subsection{Main Experiment}
To thoroughly validate the effectiveness of our hierarchical memory and the monitor-based trigger mechanism, we keep the low-level policy $\pi_{\text{e}}$ and the Planner backbone constant, varying only the \textbf{memory context} and the \textbf{planning frequency}. The comparisons are designed as follows:

\textbf{Transient Memory:} 
    A standard hierarchical VLA baseline where a memory-less Planner is invoked at fixed intervals ($n_m$) and receives only the current observation $o_t$. This represents the current paradigm of hierarchical robot foundation models, such as Hi-robot~\cite{Shi2025HiRO}.
    
\textbf{Transient Memory w/ Sentry:} 
    This variant introduces our Sentry module to trigger the Planner based on task progress. However, the Planner remains memory-less, receiving only $o_t$ upon being triggered. It evaluates the benefit of the more stabilized planning from the Sentry's gating.
    
\textbf{Flat Memory:} 
    The Planner operates at a fixed frequency and receives the 8 most recent observations, assisted by a FIFO queue of historical keyframes it previously selected. This setting tests whether unstructured, Flat Memory is sufficient for long-horizon tasks. This setting is similar to MemER \cite{Sridhar2025MemERSU}.
    
\textbf{HiMe w/o Sentry:} 
    We utilize our complete Planner's memory design but remove the sentry module. The Planner is forced to query the VLM at a fixed frequency regardless of subtask progress. This ablation validates the sentry's role in reducing computational redundancy and aligning planning steps with the dynamics of environments.
    
\textbf{HiMe (Ours):} 
    The proposed framework, where the Sentry dynamically triggers planning based on subtask progress, and the Planner utilizes the full hierarchical memory to ensure global consistency.
    
\textbf{Human High-level:} 
    A human oracle provides the correct subtask description at each step. This serves as an upper bound on task performance given the fixed capabilities of $\pi_{e}$.

\subsection{Analysis of Sentry Mechanism}
\label{sec:sentry}
We analyze the Sentry mechanism's contribution from two key dimensions: \textit{execution consistency} and \textit{observation quality}.

\begin{wraptable}[10]{r}{0.5\linewidth}
    \vspace{-1.5em}
    \centering
    \small
    \setlength{\tabcolsep}{2.5pt}
    \renewcommand{\arraystretch}{1.00}

    \caption{Hierarchical settings used in the main experiment.}
    \label{tab:main_settings}

    \begin{tabular}{@{}lll@{}}
        \toprule
        \textbf{Hierarchy Setting} &
        \textbf{Trigger} &
        \textbf{Planner Memory} \\
        \midrule
        Transient Memory &
        Periodic &
        Current observation \\
        Transient Memory w/ Sentry &
        Sentry &
        Current observation \\
        Flat Memory &
        Periodic &
        Recents + FIFO \\
        HiMe w/o Sentry &
        Periodic &
        Recents + keyframes \\
        HiMe (Ours) &
        Sentry &
        Recents + keyframes \\
        Human High-level &
        Sentry &
        Human oracle \\
        \bottomrule
    \end{tabular}
\end{wraptable}

\begin{figure*}
    \centering
    \includegraphics[width=0.9\linewidth]{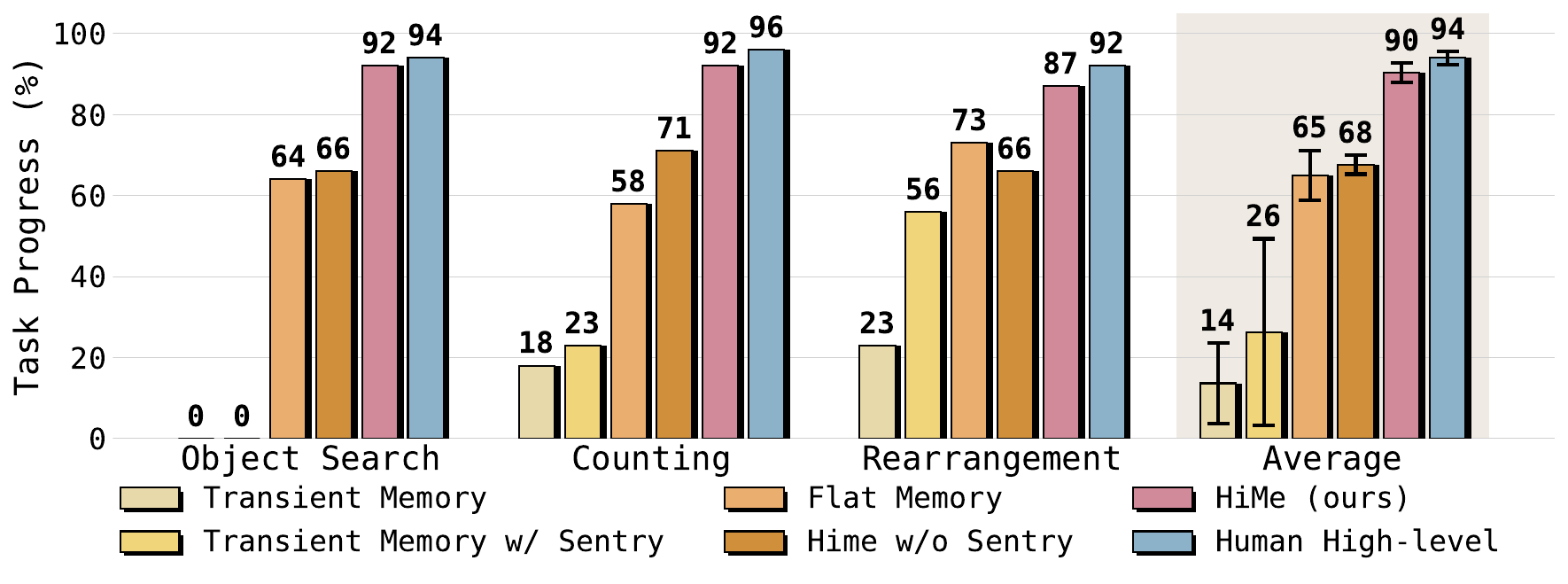}
    \caption{\textbf{Main Results.} We compare HiMe against Transient, Sentry, and Flat Memory baselines across three long-horizon tasks. HiMe significantly outperforms all baselines, achieving a 90\% average success rate. Notably, our method effectively bridges the gap between robot and the Human High-level oracle.}
    \label{fig:main}
\end{figure*}

\textit{1) Consistency via Reduced Task Switching:}
Even in a Planner with transient memory setting, the Sentry significantly boosts performance. 
In \autoref{fig:main}, we observe a substantial improvement ($14\%$ vs. $26\%$) when comparing \textit{Transient Memory} with \textit{Transient Memory w/ Sentry}. Without the Sentry, the Planner operates on a frame-by-frame style, and this high-frequency re-evaluation makes the robot hypersensitive to transient visual noise, leading to frequent, erratic switching between subtasks. The Sentry prevents the Planner from intervening until the current subtask is completed, reducing unnecessary task switching and ensuring that actions are executed coherently. It reveals that \textbf{the core value of the Sentry here is enforcing temporal consistency of Planner.}

\textit{2) Quality via Uniform Sampling vs. Recent Frames:}
The Sentry's ability to consolidate observations into high-quality working memory largely explains the gap between \textit{HiMe w/o Sentry} and \textit{HiMe}, with average task progress rising from $68\%$ to $90\%$. Whereas standard Planners rely on a narrow window of recent frames, Sentry aggregates the history of the active subtask into structured working memory. This provides a global perspective for success verification, and the consolidated memory allows the Planner to verify the entire progression, effectively reducing hallucinations and redundant replanning.

\subsection{Analysis of Planner Memory Management}
\label{sec:memory}
We validate the necessity of our hierarchical design by comparing \textit{HiMe} against memory-free baselines and flat contextual memory approaches.

\textit{(1) The Necessity of Memory:}
Long-horizon tasks require persistent state tracking. In \textit{Counting}, even with Sentry stabilization (\textit{Transient Memory w/ Sentry}), the robot still fails ($23\%$) due to the lack of persistence: without an explicit memory management, it cannot retain states (e.g., counted objects) once they leave the transient observable state.

\textit{(2) Consolidation vs. FIFO Queues:}
A critical insight from \autoref{fig:main} is the inefficiency of \textit{Flat Memory}, which functions as a contextual memory based on FIFO queues. Although it stores history, it achieves a significantly lower task progress ($65\%$) compared to \textit{HiMe} ($90\%$).
The fundamental limitation of such contextual memory is the lack of consolidation: a simple FIFO queue cannot distinguish between truly critical frames and redundant observations.
In long-horizon tasks, the limited context window is quickly flooded with noisy frames, causing earlier critical information to be discarded. \textit{HiMe} addresses this by actively consolidating memory at the subtask level. This ensures that essential high-level information is preserved regardless of the episode length, enabling high-precision retrieval with minimal Planner overhead. 

\section{Further Analysis}

\begin{figure}[t]
    \centering
    \includegraphics[width=0.9\linewidth]{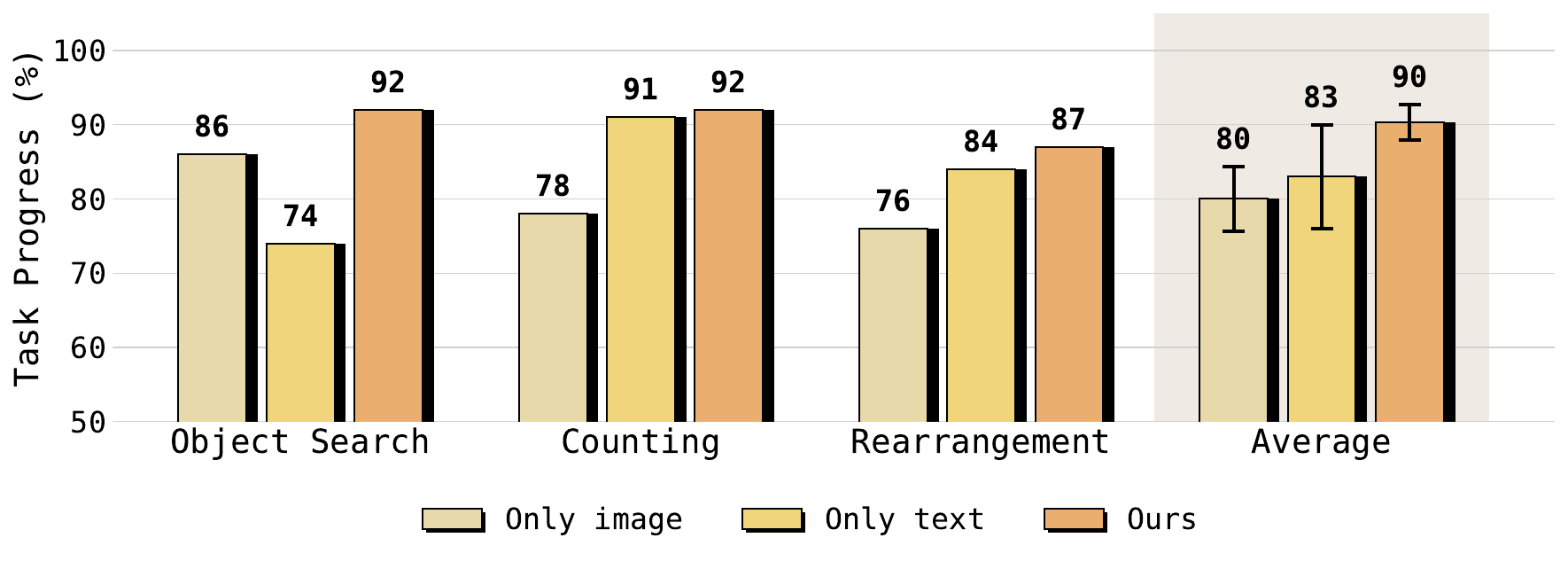}
    \caption{\textbf{Ablation of Modality.} HiMe consistently outperforms both text-only and image-only baselines across all three tasks, verifying the robustness of our cross-modal memory mechanism.}
    \label{fig:modality}
\end{figure}

\subsection{Q1: What Type of Memory Representation is Most Effective for Robotic Tasks?}
We investigate the impact of memory modality by comparing purely text memory (\textit{Only text}), purely visual memory (\textit{Only image}), and our cross-modal approach in~\autoref{fig:modality}. 

\textit{1) Spatial and Recognition Demands in Robotics:}
Our study results show that robotics tasks impose strong spatial localization and fine-grained recognition demands that caption only memory is inadequate for dynamical physical interactions. Text is a lossy compression: if perception initially misses an object or its spatial context, discarding raw visuals prevents visual re-grounding to recover details or correct errors. In \textit{Object Search}, \textit{Only image} ($86\%$) substantially outperforms \textit{Only text} ($74\%$), which shows that memory with visual evidence is significantly stronger. 

\textit{2) Text for Semantics and Logic:}
Conversely, text memory is indispensable for non-spatial reasoning. It efficiently summarizes semantics, captures logical dependencies like task sequencing, and retains user preferences. 
In tasks requiring semantic reasoning, such as \textit{Counting}, \textit{Only text} ($91\%$) outperforms \textit{Only image} ($78\%$), since these semantic clues are hard to extract from pixels on demand.

\textit{3) Superiority of Interleaved Memory:}
Our cross-modal approach achieves the best performance (Average $90\%$) by combining the spatial/recognitional fidelity of images with the semantic and logical structure of text, enabling our hierarchical memory system to perform both precise grounding and effective reasoning.

\begin{table*}[t]
\centering
\caption{Comparison of different methods. We report API calls, memory hit, and average scores across three tasks. Here, API Calls refers to the average Planner requests per subtask, and Memory Hit measures the presence of required information in memory during memory-dependent subtask.}\label{tab:comparison}
\resizebox{\textwidth}{!}{
\begin{tabular}{l ccc ccc ccc ccc}
\toprule
\textbf{Method} & \multicolumn{3}{c}{\textbf{Components}} & \multicolumn{3}{c}{\textbf{Object Search}} & \multicolumn{3}{c}{\textbf{Counting}} & \multicolumn{3}{c}{\textbf{Rearrangement}} \\
\cmidrule(lr){2-4} \cmidrule(lr){5-7} \cmidrule(lr){8-10} \cmidrule(lr){11-13}

 & \textbf{Sentry} 
 & \textbf{\begin{tabular}[b]{@{}c@{}}Memory\\Management\end{tabular}} 
 & \textbf{\begin{tabular}[b]{@{}c@{}}Memory\\Size\end{tabular}}
 & \textbf{\begin{tabular}[b]{@{}c@{}}API\\Call ($\downarrow$)\end{tabular}} 
 & \textbf{\begin{tabular}[b]{@{}c@{}}Memory\\Hit ($\uparrow$)\end{tabular}} 
 & \textbf{\begin{tabular}[b]{@{}c@{}}Avg.\\Progress ($\uparrow$)\end{tabular}} 
 & \textbf{\begin{tabular}[b]{@{}c@{}}API\\Call ($\downarrow$)\end{tabular}} 
 & \textbf{\begin{tabular}[b]{@{}c@{}}Memory\\Hit ($\uparrow$)\end{tabular}} 
 & \textbf{\begin{tabular}[b]{@{}c@{}}Avg.\\Progress ($\uparrow$)\end{tabular}} 
 & \textbf{\begin{tabular}[b]{@{}c@{}}API\\Call ($\downarrow$)\end{tabular}} 
 & \textbf{\begin{tabular}[b]{@{}c@{}}Memory\\Hit ($\uparrow$)\end{tabular}} 
 & \textbf{\begin{tabular}[b]{@{}c@{}}Avg.\\Progress ($\uparrow$)\end{tabular}} \\
\midrule

HiMe (Ours) & $\checkmark$ & $\checkmark$ & $infinite$ & \textbf{1.8} & \textbf{94\%} & \textbf{92\%} & \textbf{2.6} & \textbf{98\%} & \textbf{92\%} & \textbf{1.4} & \textbf{92\%} & \textbf{87\%} \\
Flat Memory & $\times$ & $\times$ & $recent \ 8$ & 5.4 & 68\% & 64\% & 4.8 & 61\% & 58\% & 6.2 & 76\% & 73\% \\

\bottomrule
\end{tabular}
}
\end{table*}
\subsection{Q2: Is Dynamic Memory Management Necessary?}

To assess the need for dynamic memory management, we run an ablation against two baselines: (1) \textit{No Management}, which only supports \texttt{Add}/\texttt{Retrieve} and keeps all observations as an append-only buffer; and (2) \textit{FIFO}, which caps memory at 8 entries and evicts the oldest when full.

\begin{figure}[t]
    \centering
    \includegraphics[width=0.9\linewidth]{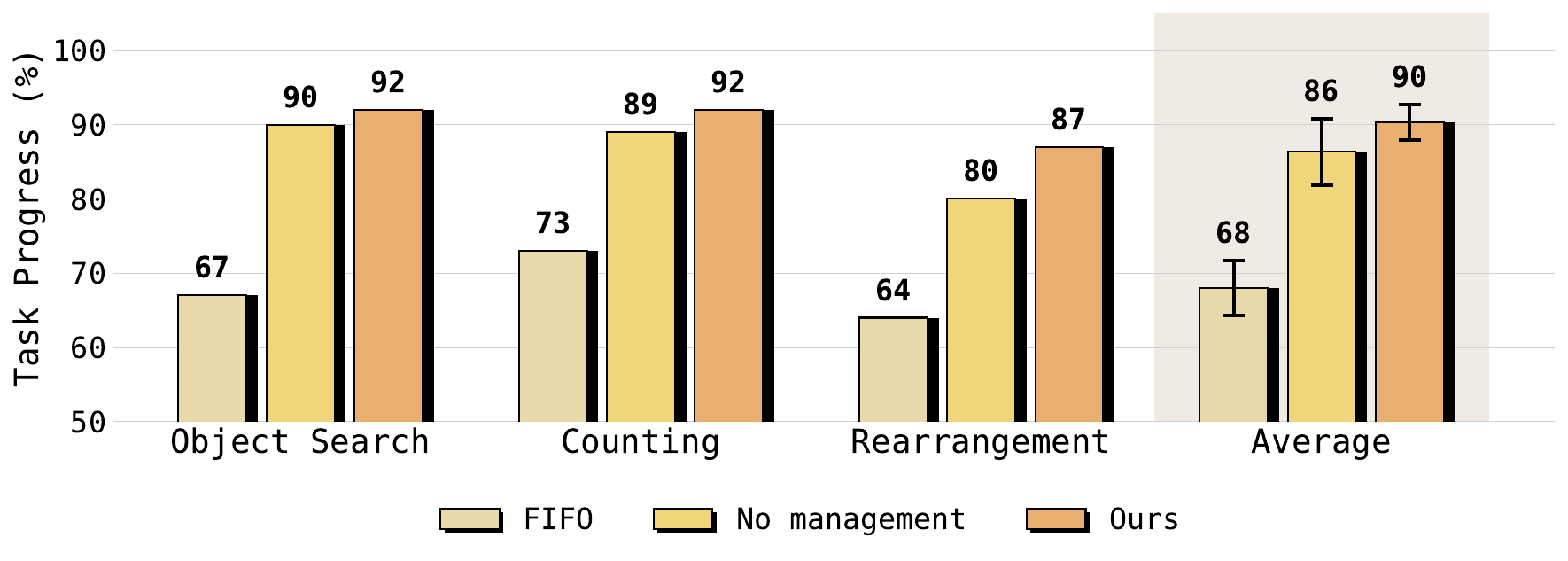}
    \caption{\textbf{Ablation of Management.} Comparison between FIFO, No management, and our approach. HiMe consistently achieves the highest task progress across all three tasks, proving the effectiveness of our memory retrieval and consolidation mechanism in long-horizon tasks.}
    \label{fig:manage}
 \end{figure}

\textit{1) The Cost of Forgetting:}
In \autoref{fig:manage}, \textit{FIFO} performs worst ($68\%$ average), far below \textit{No Management} ($86\%$). This shows that for long-horizon tasks, \textbf{early context is critical}: naive eviction can remove essential information (e.g., an object location observed early) needed for later reasoning.

\textit{2) The Value of Consistency:}
While \textit{No Management} achieves decent performance by retaining all history, it is still inferior to our full method ($86\%$ vs. $90\%$).
Although \textit{No Management} retains all history, it remains below our full method ($86\%$ vs. $90\%$) due to \textbf{redundancy and inconsistency}. Without \textit{Update}/\textit{Delete}, memory stores obsolete states (e.g., prior object locations) alongside current ones, introducing noise during \textit{Query} and potentially confusing the Planner.

\textit{3) Necessity of Active Management:}
Our method performs best by actively curating memory: \textit{Update} refreshes outdated entries to maintain consistency, and \textit{Delete} removes redundancy. Thus, storing more data is insufficient; effective robots must maintain a concise, consistent memory.

\subsection{Q3: Why Does HiMe Achieve Lower Latency and Fewer API Calls?}

We further analyze efficiency and reliability by recording total Planner invocations (average \textit{API Calls} per subtask) and the \textit{Memory Hit} rate (whether required historical context is available at query time).

Beyond the gains in ~\autoref{sec:sentry} and \autoref{sec:memory}, ~\autoref{tab:comparison} shows a clear efficiency advantage: \textit{HiMe} reduces API calls by $\sim3\times$ (e.g., 5.4 to 1.8 in Object Search) over Flat Memory. Since VLM inference dominates latency, fewer calls directly yield faster execution. 
This efficiency stems from two structural advantages over the Flat memory:

\textit{1) Sentry Reduces Invocation Frequency:}
In Flat memory methods, the Planner is often queried every step to interpret the visual buffer, placing the expensive LLM in a high-frequency control loop. Instead, after the Planner specifies a subtask, the Sentry handles dense completion checking and wakes the Planner only when necessary. Thus, Planner involvement shifts from \textit{step-by-step} to sparse \textit{subtask-level} calls.

\begin{wrapfigure}{r}{0.38\linewidth}
    \centering
    \vspace{-15pt}
    \includegraphics[width=\linewidth]{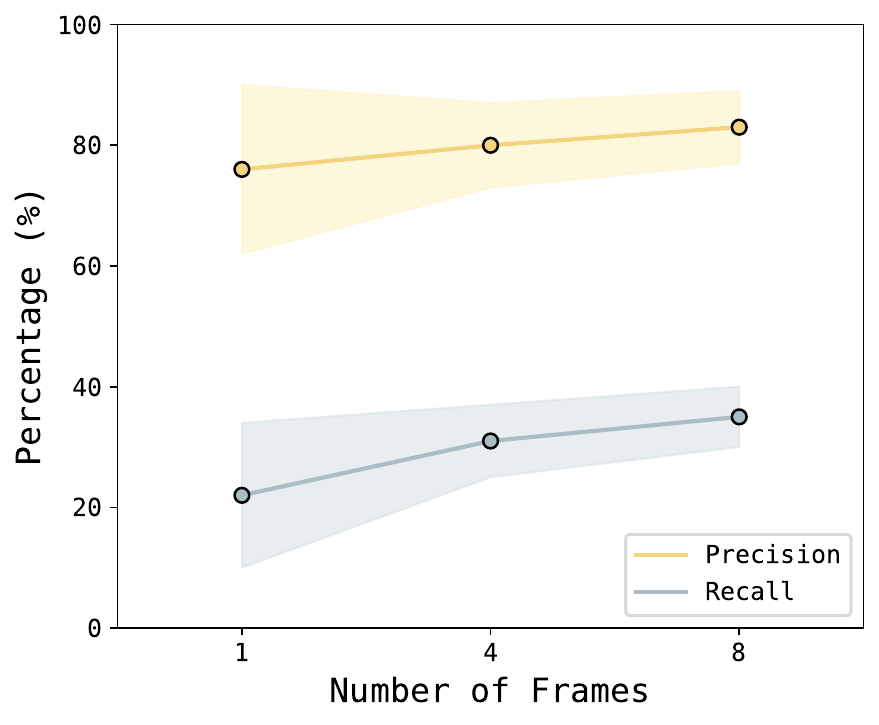}
    \caption{Increasing the number of recent frames provided to the Sentry leads to a consistent improvement in both precision and recall. This indicates that a longer temporal context is beneficial for accurate subtask monitoring.}
    \vspace{-40pt}
    \label{fig:sentry_ablation}
\end{wrapfigure}

\textit{2) Infinite Memory vs. FIFO Forgetting:}
Flat Memory uses a FIFO buffer (\autoref{tab:main_settings}), forming a sliding window that forgets early observations. The resulting low memory hit rate ($68\%$ vs. $94\%$) forces redundant exploration when needed context has been evicted. In contrast, HiMe maintains an infinite structured memory, avoiding this forgetting and enabling immediate retrieval without physical re-exploration.

\subsection{Q4: What Affects Sentry's Termination Behavior?}

To understand the Sentry's decision-making boundary, we analyzed its performance on subtask completion detection under different working memory sizes (window length $N$). \autoref{fig:sentry_ablation} presents the Precision and Recall, where Subtask ``Done'' is treated as the positive class.

\textit{1) Benefits of Temporal Context:}
As shown in the \autoref{fig:sentry_ablation}, \mbox{increasing} the context window from 1 to 8 frames improves both Precision (from $\sim76\%$ to $\sim82\%$) and Recall (from $\sim22\%$ to $\sim35\%$).
A single frame often lacks sufficient information to distinguish between a temporary pause and true \mbox{completion}.
By observing a sequence of 8 frames (consistent with our sentry‘s working memory), the Sentry can leverage temporal cues to make more reliable judgments.

\textit{2) The "Conservative" Nature of Sentry:}
The significant gap between high Precision and low Recall reveals the Sentry's inherent ``conservative'' nature.
The system exhibits a strong bias towards the incomplete state, preferring to continue execution unless overwhelmingly confident that the goal is met.
While this minimizes premature stops (high Precision), the low Recall creates a ``missing signal'' problem where the agent might overshoot its target or enter infinite loops.
Since the Sentry is prone to False Negatives (missing the ``Done'' event), we design a fixed-interval Planner fallback. It ensures that execution loops are eventually broken even when the Sentry fails to trigger.

\section{Conclusion}
In this work, we presented HiMe, a novel hierarchical embodied memory framework that resolves the fundamental frequency-competence paradox in long-horizon Vision-Language-Action control. By decoupling embodied intelligence into a high-frequency Executor, a progress-aware Sentry, and a strategic Planner, we mirror the multi-store structure of human cognition to balance real-time responsiveness with deep reasoning. Our introduction of cross-modal semantic schemata and active management mechanisms (Add, Update, Delete) transforms robotic memory from a passive observation buffer into a dynamic, self-evolving knowledge system. Extensive experiments demonstrate that HiMe achieves 90\% average success rate—effectively bridging the gap to human-level performance. 
We believe that this shift from flat, transient policies to hierarchical, structured memory is a vital step toward developing truly autonomous and adaptive robotic agents.

\section*{Acknowledgements}
This work was supported by the National Natural Science Foundation of China (No. 62521004). 
We thank the incredible SII \mbox{MakerClub} for their generous supply of essential materials and invaluable technical assistance throughout this project. 
We are also deeply grateful to Chunbiao Feng and Hongbo Tang for their pivotal technical support in hardware setup and control implementation.

\section*{Impact Statement}
This paper presents work where the goal is to advance the fields of machine learning and robotics. There are many potential societal consequences of our work, none of which we feel must be specifically highlighted here.

\section*{Limitations}
The paper has several limitations that suggest directions for future work. Our evaluation is primarily conducted on real-robot tasks without a matched standard simulation benchmark. While this setting better captures memory-intensive long-horizon behavior under real-world dynamics, it limits controlled comparison and benchmark-level reproducibility. In addition, the current tasks mainly rely on relatively simple manipulation primitives such as pick-and-place, and broader validation on more diverse and dexterous skills remains necessary. Although we provide additional analysis on memory scaling and open-source Planner substitution, the current study does not yet fully characterize performance under very long horizons, larger environments, or extended deployment over time.

\clearpage
\bibliographystyle{unsrtnat}
\bibliography{main}

\clearpage
\beginappendix





\section{Task Design and Metrics}

Our real-robot evaluation consists of three types of structured tasks: \textbf{Object Search}, \textbf{Counting}, and \textbf{Rearrangement}. These tasks are designed to cover the aspects of long-horizon embodied decision-making, including inspection-based memory acquisition, persistent state tracking, dynamic memory revision, and preference-aware planning.

\paragraph{Tasks.}
Table~\ref{tab:task_design} summarizes the task settings and their corresponding memory challenges.

\begin{table}[H]
\centering
\small
\setlength{\tabcolsep}{5pt}
\caption{Real-robot tasks in our evaluation.}
\begin{tabular}{p{2.3cm} p{5.2cm} p{5.0cm}}
\toprule
\textbf{Task} & \textbf{Setting} & \textbf{Core Memory Challenge} \\
\midrule
\textbf{Object Search} & 
Three boxes with unseen internal contents and visible objects on the table. The robot must actively inspect the boxes to infer their contents and then place visible objects accordingly. During execution, the environment may change, requiring memory revision. & 
Memory formation under partial observability through \textbf{active inspection}, together with \textbf{continuous memory updates} in changing environments. \\
\midrule
\textbf{Counting} & 
Multiple ingredient plates and a recipe specifying required proportions. The agent must first read the recipe and then repeatedly pick ingredients across multiple steps. & 
Persistent semantic memory and \textbf{cumulative progress tracking} over long horizons. \\
\midrule
\textbf{Rearrangement} & 
Multiple objects and a container. The agent must first collect all objects while remembering their original locations, and later restore them to their original positions. & 
Long-horizon memory retention and consistency over extended execution. \\
\bottomrule
\end{tabular}
\label{tab:task_design}
\end{table}

Although these tasks share the same hierarchical control framework, they stress different forms of memory usage. \textbf{Object Search} emphasizes active perception and belief updating under partial observability, since the robot cannot determine box contents without inspection and must revise memory when the environment changes. \textbf{Counting} focuses on persistent semantic memory and progress monitoring, as the robot must retain the recipe requirements and keep track of cumulative ingredient collection across repeated action cycles. \textbf{Rearrangement} requires longer-term spatial memory, because the robot must remember the original object layout during collection and later use this stored information to restore the scene.

Across all task families, we additionally introduce \textbf{user preference constraints}, such as preferred objects, disliked ingredients, or placement rules. These constraints require the agent not only to store environment states, but also to maintain and use user-specific information during planning and execution. As a result, the system must integrate memory with high-level reasoning and dynamically adjust its plan when new observations or updated preferences become relevant.

\paragraph{Action steps.}
To further characterize the difficulty of these real-robot tasks, we report the number of subtasks and the approximate total number of action steps required by each task in Table~\ref{tab:task_complexity}. These statistics reflect that all three tasks require multi-stage execution over extended horizons, rather than single-step or short reactive behaviors.

\begin{table}[H]
\centering
\small
\setlength{\tabcolsep}{6pt}
\caption{Action steps across 3 tasks.}
\begin{tabular}{lcc}
\toprule
\textbf{Task} & \textbf{Number of Subtasks} & \textbf{Total Action Steps} \\
\midrule
Object Search & 6 & $\sim$1450 \\
Counting & 7 & $\sim$1200 \\
Rearrangement & 6 & $\sim$1035 \\
\bottomrule
\end{tabular}
\label{tab:task_complexity}
\end{table}

\paragraph{Evaluation.}
\label{app:eval}
Each task is evaluated over \textbf{20 trials per method}. We measure performance using the metrics summarized in Table~\ref{tab:task_metrics}. These metrics are designed to capture three complementary aspects of system performance. \textbf{Task Progress} measures execution quality at the task level, indicating whether the agent can successfully complete the required long-horizon objectives. \textbf{API Calls} measures reasoning frequency and therefore serves as a proxy for computational cost and system efficiency. \textbf{Memory Hit Rate} directly evaluates whether the maintained memory contains the information needed for effective replanning and decision-making when the Planner is invoked. Together, these metrics provide a joint assessment of task completion quality, computational efficiency, and memory effectiveness.

\begin{table}[H]
\centering
\small
\setlength{\tabcolsep}{6pt}
\caption{Evaluation metrics.}
\begin{tabular}{p{3cm} p{9.2cm}}
\toprule
\textbf{Metric} & \textbf{Description} \\
\midrule
\textbf{Task Progress} & 
Each task is decomposed into multiple subtasks. Task Progress measures the proportion of completed subtasks, reflecting the overall task completion level. \\
\midrule
\textbf{API Calls} & 
The average number of Planner invocations per subtask, reflecting reasoning frequency and computational cost. \\
\midrule
\textbf{Memory Hit Rate} & 
Whether the required information is present in memory when the Planner is invoked, reported as the probability of successful memory retrieval. \\
\bottomrule
\end{tabular}
\label{tab:task_metrics}
\end{table}

\section{Model Initialization, Training, and Deployment}
\label{sec:finetuning_hyperparams}

We provide the initialization, training, and deployment details of the three-layer HiMe system, including the low-level Executor, the high-level Planner, and the Sentry module.

\paragraph{Model setup.}
HiMe consists of three components: a low-level Executor for action generation, a high-level Planner for memory update and subtask decomposition, and a Sentry for execution monitoring and replanning trigger prediction. The overall model setup is summarized in Table~\ref{tab:model_setup}.

\begin{table}[H]
\centering
\caption{Model initialization and deployment setup of HiMe.}
\label{tab:model_setup}
\begin{tabular}{lll}
\toprule
\textbf{Module} & \textbf{Model} & \textbf{Role} \\
\midrule
Planner & GPT-4o / Qwen3-VL-30B & Memory update and replanning \\
Sentry & Qwen3-VL-8B & Execution monitoring and trigger prediction \\
Executor & $\pi_{0.5}$ & Low-level action generation \\
\bottomrule
\end{tabular}
\end{table}

The \textbf{Executor} is initialized from the public $\pi_{0.5}$ checkpoint trained on the DROID dataset~\citep{Khazatsky2024DROIDAL}, and is further fine-tuned on our task-specific real-robot demonstrations. The \textbf{Planner} is instantiated with GPT-4o in the main experiments. To improve reproducibility, we additionally replace the Planner with an open-source VLM, Qwen3-VL-30B, in supplementary experiments under the same evaluation protocol. The \textbf{Sentry} is implemented with Qwen3-VL-8B and deployed locally for online monitoring during execution.

\paragraph{Robot observations and control interface.}
The Executor takes as input two RGB observations, including a third-person view and a wrist camera view, together with the robot state consisting of the end-effector pose and gripper state. It predicts low-level action chunks for control. In our implementation, the Executor predicts 10 actions at a time, corresponding to an action horizon of 10. The robot platform is a WidowX-250 manipulator with a parallel gripper. The overall control loop runs at 10\,Hz.

\paragraph{Low-level policy training.}
The Executor is fine-tuned on task-specific real-robot demonstrations collected on the WidowX-250 platform. Our training set contains 60 demonstrations in total, including 50 standard demonstrations and 10 additional corner-case demonstrations. We follow the standard OpenPI training recipe. The main hyperparameters are listed in Table~\ref{tab:low_level_hyperparams}.

\begin{table}[H]
\centering
\caption{Hyperparameters for low-level Executor ($\pi_{0.5}$) fine-tuning.}
\label{tab:low_level_hyperparams}
\begin{tabular}{ll}
\toprule
\textbf{Hyperparameter} & \textbf{Value} \\
\midrule
Optimizer & AdamW \\
$\beta_1$ & 0.9 \\
$\beta_2$ & 0.999 \\
Weight Decay & 0.1 \\
Gradient Clip Norm & 1.0 \\
LR Schedule & Cosine decay \\
Warmup Steps & 10k \\
Peak Learning Rate & $5 \times 10^{-5}$ \\
Batch Size & 256 \\
EMA Decay & 0.999 \\
Training Steps & 30k \\
Action Horizon & 10 \\
Demonstrations & 60 (50 regular + 10 corner cases) \\
\bottomrule
\end{tabular}
\end{table}

\paragraph{Planner and Sentry deployment.}
The Planner is invoked only when replanning is required, rather than at every control step. In the main experiments, the Planner is instantiated with GPT-4o through the OpenAI API. For local deployment experiments, we serve Qwen3-VL models with vLLM through an OpenAI-compatible API interface. The detailed deployment configuration is summarized in Table~\ref{tab:vlm_deployment}.

\begin{table}[H]
\centering
\caption{Deployment configuration for locally served VLM modules. Unless otherwise specified, all other settings use default vLLM configurations.}
\label{tab:vlm_deployment}
\begin{tabular}{lllll}
\toprule
\textbf{Module} & \textbf{Model} & \textbf{GPU Setup} & \textbf{Temperature} & \textbf{Max Model Len} \\
\midrule
Sentry & Qwen3-VL-8B & 1$\times$ H100 (80GB) & 0.6 & 8192 \\
Planner (supp.) & Qwen3-VL-30B & 2$\times$ H100 (80GB) & 0.6 & 16384 \\
\bottomrule
\end{tabular}
\end{table}

\paragraph{Inference setup.}
Both local models are served with vLLM using an OpenAI-compatible endpoint and integrated into the online real-robot system through standard API-based invocation.

\begin{figure}[H]
    \centering
    \includegraphics[width=0.9\linewidth]{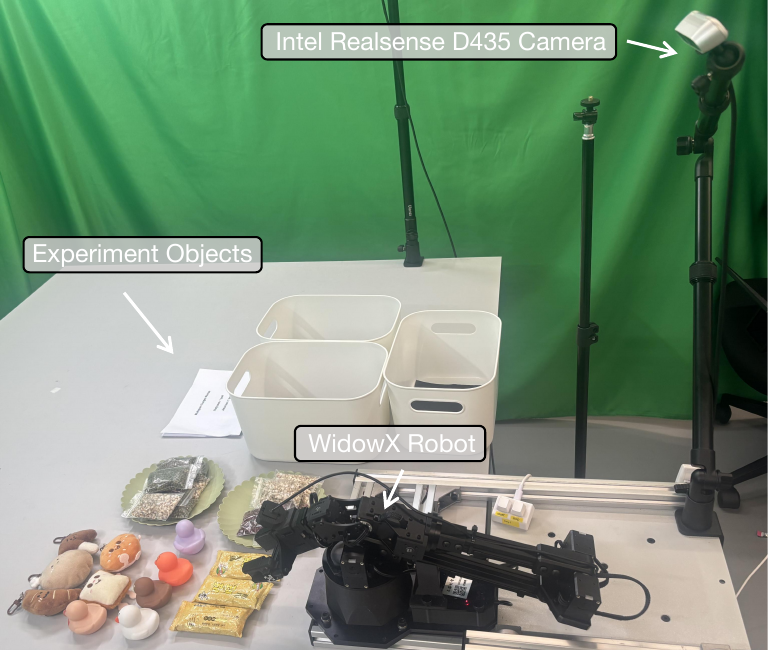}
    \caption{Real-world experimental setup. The system consists of a WidowX-250 robot arm, a front-facing Intel RealSense D435 camera, and a tabletop workspace containing task objects and storage boxes.}
    \label{fig:real_robot_setup}
\end{figure}

\paragraph{Real-robot setup.}
All task-specific demonstrations are collected on a real WidowX-250 platform under a fixed tabletop setup, as shown in Figure~\ref{fig:real_robot_setup}. The setup includes a front-facing Intel RealSense D435 camera for third-person observation, a wrist-mounted camera for close-range manipulation feedback, a set of task objects placed in the shared workspace, and multiple containers used for long-horizon manipulation tasks. Demonstrations are collected through leader-follower teleoperation, and all robot control and sensor streams are integrated through ROS.

During data collection, both the external camera and the wrist camera record RGB observations at 25\,Hz. Each trajectory is synchronized with robot proprioceptive states, including end-effector pose, joint states, and gripper state. After collection, raw trajectories are processed through a standardized preprocessing pipeline. All images are resized to $224 \times 224$, and the trajectories are temporally subsampled to 10\,Hz. The processed demonstrations are then converted into the LeRobot format for downstream training.

\section{Additional Experiments}
\label{sec:additional_experiments}

We provide additional experiments to further analyze controlled memory management, Planner generalization with open-source VLMs, memory scalability over extended horizons, and system latency.

\subsection{Controlled Memory Ablation}
\label{sec:controlled_memory_ablation}

We first conduct a controlled memory ablation on the \textbf{Rearrangement} task to separate the effect of memory management strategy from that of memory capacity. In the main setting, HiMe uses unconstrained active memory, while the FIFO baseline is evaluated with a fixed memory budget. Although the FIFO budget is chosen to match the average memory size of the unlimited active baseline, the two settings still differ in whether memory is explicitly constrained. To provide a cleaner comparison, we additionally evaluate a limited-memory version of active management under the same fixed budget.

Specifically, we compare three variants: \textbf{Unlimited Active Management}, corresponding to the original HiMe setting with unconstrained memory; \textbf{Limited Active Management}, which uses the same Add / Update / Delete mechanism under a fixed budget of 8 entries; and \textbf{FIFO (Limited)}, which uses the same budget with FIFO eviction.

\begin{table}[H]
\centering
\caption{Controlled memory in Rearrangement.}
\label{tab:controlled_memory_ablation}
\begin{tabular}{lcc}
\toprule
\textbf{Method} & \textbf{Memory Budget} & \textbf{Task Progress (\%)} \\
\midrule
Unlimited Active Management & Unbounded & 87.0 \\
Limited Active Management & Fixed (8 entries) & 80.0 \\
FIFO (Limited) & Fixed (8 entries) & 64.0 \\
\bottomrule
\end{tabular}
\end{table}

Performance decreases under constrained memory, but active management remains substantially stronger than FIFO under the same capacity. This result indicates that the benefit of HiMe is not explained by larger memory alone, and that explicit Update / Delete operations remain useful beyond simple eviction.

\subsection{Open-Source Planner Evaluation}
\label{sec:opensource_planner_eval}

\begin{figure}[H]
    \centering
    \includegraphics[width=\linewidth]{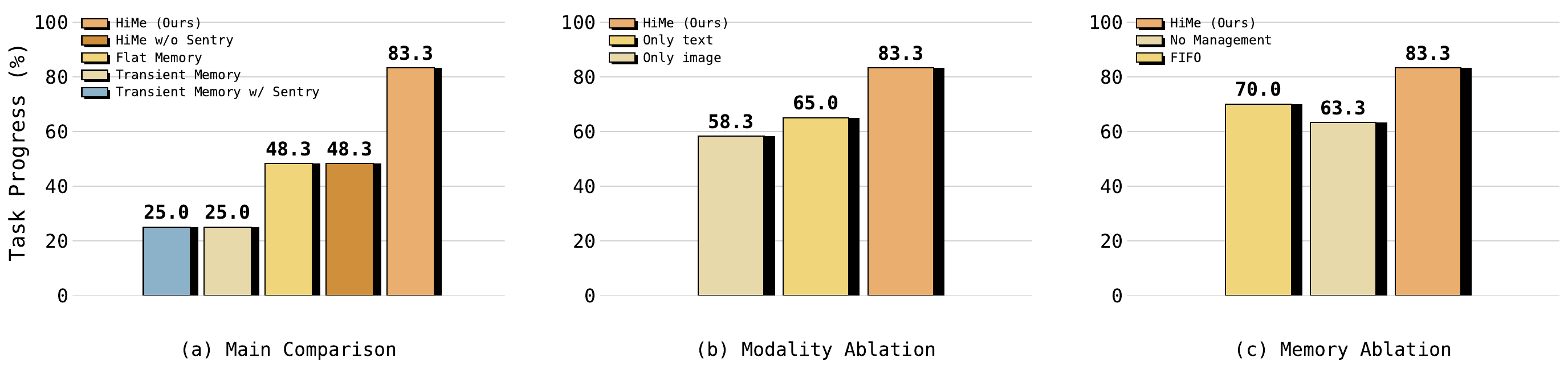}
    \caption{Additional experiments on the \textbf{Rearrangement} task with \textbf{Qwen3-VL-30B} as the Planner. (a) Main comparison with the corresponding baselines. (b) Modality ablation. (c) Memory ablation. The overall trends are consistent with the main experiments, and HiMe remains the best-performing variant under the open-source Planner.}
    \label{fig:qwen_appendix_more_exp}
\end{figure}

To improve reproducibility and examine Planner generalization, we replace the main Planner with the open-source VLM \textbf{Qwen3-VL-30B} and evaluate the system on the \textbf{Rearrangement} task under the same protocol. The results are shown in Fig.~\ref{fig:qwen_appendix_more_exp}.

Fig.~\ref{fig:qwen_appendix_more_exp}(a) reports the main comparison with the corresponding baselines. Fig.~\ref{fig:qwen_appendix_more_exp}(b) presents the modality ablation, comparing the full system with text-only and image-only memory. Fig.~\ref{fig:qwen_appendix_more_exp}(c) shows the memory ablation under the same open-source Planner. Overall, the trends are consistent with the main experiments: HiMe remains the best-performing variant with Qwen3-VL-30B, indicating that the benefits of structured memory and sentry-based coordination are robust to the choice of Planner.

\subsection{Memory Scalability}
\label{sec:memory_scalability}

We further analyze memory scalability using an extended multi-round version of \textbf{Object Search}. In each round, the robot must inspect boxes with initially unknown contents and place the visible table object accordingly. At the beginning of each new round, the box contents are changed, requiring the agent to re-inspect the scene and revise memory before acting. As the number of rounds increases, both the task horizon and the number of subtasks increase.

We report the average \textbf{Memory Size} and \textbf{Task Progress} across different numbers of rounds.

\begin{table}[H]
\centering
\caption{Memory scalability in extended multi-round Object Search.}
\label{tab:memory_scalability}
\begin{tabular}{cccc}
\toprule
\textbf{Rounds} & \textbf{Subtasks} & \textbf{Memory Size} & \textbf{Task Progress (\%)} \\
\midrule
1 & 4  & 5.5  & 92.5 \\
2 & 8  & 9.9  & 76.3 \\
3 & 12 & 14.0 & 66.7 \\
\bottomrule
\end{tabular}
\end{table}

As the horizon increases, memory size grows while task performance declines. This result suggests that the main challenge in long-horizon settings is not only storing more information, but also maintaining correct and up-to-date memory as the environment evolves.

\subsection{System Latency}
\label{sec:system_latency}

We additionally measure Planner latency under different deployments. Each Planner call consists of two stages: memory query / retrieval, and plan generation together with memory update. We report end-to-end latency over complete Planner calls.

\begin{table}[H]
\centering
\caption{End-to-end Planner latency under different model deployments.}
\label{tab:system_latency}
\begin{tabular}{lcccc}
\toprule
\textbf{Model} & \textbf{P50 (s)} & \textbf{P90 (s)} & \textbf{P99 (s)} & \textbf{Avg. Completion Tokens} \\
\midrule
GPT-4o (API) & 38.59 & 49.30 & 57.28 & 517.9 \\
Qwen3-VL-30B (local) & 6.83 & 7.93 & 8.65 & 532.8 \\
Qwen3-VL-8B (local) & 6.10 & 7.08 & 11.34 & 399.0 \\
\bottomrule
\end{tabular}
\end{table}

Latency differences are mainly associated with deployment mode, i.e., remote API versus local serving, rather than token scale alone. In practice, HiMe is compatible with different Planner backends, and local deployment substantially reduces reasoning latency. Since Planner calls are triggered only when necessary, this reasoning latency remains decoupled from the high-frequency low-level control loop.

\newpage
\section{Prompts Details}

\begin{tcolorbox}[colback=black!2, colframe=black!45, title=Sentry Prompt]
You are the \textbf{EXECUTION OBSERVER} of an embodied agent.

\textbf{Primary input modality:}
\begin{itemize}[leftmargin=*]
    \item \textbf{Plan list:}
    \begin{itemize}
        \item The model’s current decomposition of the task into subtasks.
        \item Each subtask line includes its execution status (e.g., [done], [current], [pending]).
    \end{itemize}
    
    \item \textbf{Combined images of recent frames:}
    \begin{itemize}
        \item \textbf{CRITICAL LAYOUT INFO:} Each image is a composite.
        \item LEFT HALF: Wrist Camera View (Close-up, attached to the gripper).
        \item RIGHT HALF: Third-Person View (Global view of the robot and environment).
    \end{itemize}
\end{itemize}

Your \textbf{ONLY} job is to judge whether the \textbf{CURRENT} subtask has been completed.

\textbf{Judgment Criteria:}
\begin{itemize}[leftmargin=*]
    \item \textbf{Inspect:} done when the robot arm is above or extended into the box, and the wrist view shows the content and the black background of the box.
    
    \item \textbf{Pick \& Place:} done when the object has been successfully released at the destination or you observe that the robot arm is above a box and see its black background.
    
    \item \textbf{Reset:} done when the robot arm is in the home pose.
    \begin{itemize}
        \item The robot arm home pose is: in both images, the arm is aligned parallel to what appears to be a rail or track. The end effector (or gripper) is open and facing downwards towards the table.
    \end{itemize}
    
    \item The interior bottom surface of the box is lined with a black background to enhance contrast for detection.
\end{itemize}

\noindent\rule{\linewidth}{0.4pt} 

\textbf{\# REQUIRED OUTPUT FORMAT}

You \textbf{MUST} output exactly this XML structure:

\begin{verbatim}
<status>
<!-- EXACTLY one of:
        done
        not_done
-->
</status>
\end{verbatim}

\end{tcolorbox}

\begin{tcolorbox}[
    enhanced,                   
    colback=white,              
    colframe=blue!50!black,     
    coltext=black,              
    title=Planner Prompt,
    fonttitle=\bfseries,        
    breakable,                  
    arc=2mm                     
]

You are the \textbf{PLANNER} of an embodied agent.

\textbf{Primary input modalities:}
\begin{itemize}[leftmargin=*]
    \item \textbf{User instruction:} The user's high-level instruction for the agent.
    \item \textbf{Plan list:} The model's current decomposition of the task into subtasks. Each subtask line includes its execution status (e.g., [done], [current], [pending]).
    \item \textbf{Combined images during subtask execution:}
    \begin{itemize}
        \item \textbf{IMAGE LAYOUT \& USAGE GUIDE:}
        \item \textbf{LEFT HALF (Wrist View):} Egocentric view attached to the gripper. Usage: Identify specific objects, read text, verify grasping.
        \item \textbf{RIGHT HALF (Third-Person View):} Global view. Usage: Spatial context, locating containers, tracking arm position.
        \item \textbf{NOTE:} Integrate information from both. Use Right view to navigate, Left view to inspect.
    \end{itemize}
\end{itemize}

You have access to an external \textbf{MEMORY} module through a structured CRUD interface.
MEMORY stores records with fields: \texttt{id}, \texttt{tags}, \texttt{data} (type, value), \texttt{image\_path}.

\noindent\rule{\linewidth}{0.4pt}

\textbf{\# TAG DESIGN PRINCIPLES}

Tags are the PRIMARY mechanism for memory retrieval.

\begin{enumerate}[leftmargin=*]
    \item \textbf{OBJECT TAGS:} \texttt{toy\_duck}, \texttt{snack\_package}, \texttt{left\_box}.
    \item \textbf{LOCATION TAGS:} \texttt{table}, \texttt{shelf}, \texttt{inside\_left\_box}.
    \item \textbf{USER\_PREFERENCE TAGS:} \texttt{user}, \texttt{lily}, \texttt{likes}, \texttt{prefers}.
\end{enumerate}

\textbf{Tag Usage Rules:} Use 2-5 tags per record. Update tags when object state changes.
\textbf{Tag Query Strategy:} PREFER tag-based queries. Query one tag every time.

\noindent\rule{\linewidth}{0.4pt}

\textbf{\# MEMORY CRUD PROTOCOL}

You interact with memory only via structured XML operations.

\textbf{(1) QUERY (READ)}
\begin{verbatim}
<operation>
  <type>QUERY</type>
  <query>search key tags</query>
  <reason>why you search this</reason>
</operation>
\end{verbatim}

\textbf{(2) CREATE}
\begin{verbatim}
<operation>
  <type>CREATE</type>
  <tags>...</tags>
  <text>concise description or fact</text>
  <image_path>index</image_path>
  <reason>why this new memory is necessary</reason>
</operation>
\end{verbatim}

\textbf{(3) UPDATE}
\begin{verbatim}
<operation>
  <type>UPDATE</type>
  <id>record_id</id>
  <tags>...</tags>
  <text>updated description</text>
  <image_path>2</image_path>
  <reason>why this existing record must be changed</reason>
</operation>
\end{verbatim}

\textbf{(4) DELETE}
\begin{verbatim}
<operation>
  <type>DELETE</type>
  <id>...</id>
  <reason>why this record is no longer useful</reason>
</operation>
\end{verbatim}

\noindent\rule{\linewidth}{0.4pt}

\textbf{\# EVIDENCE RELIABILITY RULES (NO ASSUMPTIONS)}
\begin{itemize}[leftmargin=*]
    \item Memory must contain only verifiable facts derived from explicit user instructions or direct visual evidence.
    \item \textbf{Do NOT CREATE FAKE memory.} Prefer planning an inspection step.
    \item UPDATE is for correcting/refining without changing time meaning.
    \item When state changes, \textbf{CREATE} a new record, UPDATE the old one to label as past.
    \item DELETE sparingly (only for duplicates or errors).
\end{itemize}

\noindent\rule{\linewidth}{0.4pt}

\textbf{\# MEMORY USAGE POLICY (TWO-TURN INTERACTION)}
\begin{itemize}[leftmargin=*]
    \item \textbf{Turn 1 (QUERY-only):} Issue QUERY operations. NO Create/Update/Delete. Plan list must be empty.
    \item \textbf{Turn 2 (REFINE + CRUD):} Issue Create/Update/Delete based on query results. Output finalized \texttt{<plan\_list>}.
\end{itemize}

\noindent\rule{\linewidth}{0.4pt}

\textbf{\# PLAN LIST FORMAT (FINAL OUTPUT ONLY, TURN 2)}
Plain text, one subtask per line.
\begin{itemize}
    \item \texttt{[done]}: subtask completed.
    \item \texttt{[current]}: the single subtask to execute NEXT.
    \item \texttt{[pending]}: future subtasks.
\end{itemize}
Annotation: You MAY include a short purpose in parentheses, e.g., \texttt{[pending] inspect recipe (check what does it need)}.

\noindent\rule{\linewidth}{0.4pt}

\textbf{\# PLAN LIST ACTION FORMAT}
\begin{enumerate}[leftmargin=*]
    \item \textbf{Inspection action:} Main verb "inspect". Use when information is missing.
    \item \textbf{Pick-and-place action:} "pick up \texttt{<object>} \texttt{<source>} and place it to \texttt{<target>}".
\end{enumerate}

\noindent\rule{\linewidth}{0.4pt}

\textbf{\# REQUIRED OUTPUT FORMAT FOR EACH TURN}

\begin{verbatim}
<summary>
  <!-- Concise reasoning summarizing observations and memory interaction -->
</summary>

<memory_operations>
  <!-- Turn 1: QUERY only. Turn 2: CRUD operations. -->
</memory_operations>

<plan_list>
  <!-- Turn 1: EMPTY.
       Turn 2: Finalized plan list.
       [done] ...
       [current] ...
       [pending] ...
  -->
</plan_list>
\end{verbatim}

\end{tcolorbox}


\end{document}